\documentclass[11pt,a4paper]{article}
\usepackage[hyperref]{emnlp2020}
\usepackage{times}
\usepackage{latexsym}
\usepackage{balance}

\usepackage{microtype}

\aclfinalcopy % Uncomment this line for the final submission
\title{Dartmouth CS at WNUT-2020 Task 2: Informative COVID-19 Tweet Classification Using BERT}
\author{Dylan Whang \\
  Dartmouth College \\
  Hanover, NH \\
  \texttt{dylanmwhang@gmail.com} \\\And
  Soroush Vosoughi \\
  Dartmouth College \\
  Hanover, NH \\
  \texttt{soroush@dartmouth.edu} \\}

\date{}

\begin{document}
\maketitle
\begin{abstract}
We describe the systems developed for the WNUT-2020 shared task 2, identification of informative COVID-19 English Tweets. BERT is a highly performant model for Natural Language Processing tasks. We increased BERT’s performance in this classification task by fine-tuning BERT and concatenating its embeddings with Tweet-specific features and training a Support Vector Machine (SVM) for classification (henceforth called BERT+). We compared its performance to a suite of machine learning models. We used a Twitter specific data cleaning pipeline and word-level TF-IDF to extract features for the non-BERT models. BERT+ was the top performing model with an F1-score of 0.8713.
\end{abstract}

\section{Introduction}
In an effort to aid automated the development of COVID-19 related monitoring systems, the WNUT-2020 shared task 2: Identification of informative COVID-19 English Tweets tasked participants with developing systems to automatically classify Tweets as \verb|INFORMATIVE| or \verb|UNINFORMATIVE|. The WNUT task organizers have constructed and provided a data set of 10,000 Tweets related to Covid-19 for this task. \citep{covid19tweet}.

For this language processing task, we used Google's Bidirectional Encoder Representations from Transformers (BERT) to achieve performant results. BERT uses the now ubiquitous Transformer neural network architecture as explained in depth in the article, ``Attention is All You Need,'' \cite{vaswani2017attention} and garnered acclaim for obtaining new state-of-the-art results on eleven natural language processing tasks, including pushing the GLUE score to 80.5\%. \cite{devlin-etal-2019-bert} 

To optimize BERT for this task, we fine-tuned the \verb|BERT-large-uncased| pretrained language model on the WNUT-2020 shared task 2 data set. We then further improved performance by concatenating the fine-tuned BERT embedding vectors with Tweet-specific features and using a Support Vector Machine (SVM) for classification (BERT+).

To benchmark the performance of the BERT+ model, we compared its performance to five traditional classifiers. We also developed a preprocessing pipeline for data cleaning and used Text Frequency Inverse Document Frequency (TF-IDF) to extract features for the traditional classifiers.

\subsection{Pretrained BERT model}
We used the \verb|BERT-large-uncased| pretrained language model. This BERT model contains an encoder with 24 Transformer blocks, 16 self-attention heads, with the hidden size of 1024. 

BERT generates its pretrained word and sentence level embeddings by using two objectives: Masked Language Modeling (MLM) and Next Sentence Prediction (NSP). 

During pretraining, BERT utilizes MLM by first selecting 15\% of the inputted tokens for potential masking. Our of this 15\%, 80\% are replaced with the \verb|[MASK]| Token, 10\% are replaced by a randomly selected word, and the remaining 10\% are not manipulated. The MLM Objective is a cross-entropy loss on predicting the masked tokens. 

BERT also uses NSP in its pretraining. The NSP objective is a binary classification loss for predicting if two sequences follow each other. NSP uses an equal proportion of consecutive sentences for the text corpus as positive examples and randomly paired sentences as negative examples. \citep{liu2020roberta}

The ``BERT for sequence classification'' model utilizes the special \verb|[CLS]| classifier token as the first token in every sequence. This token contains the classification embedding of the sequence. BERT uses the final hidden state of the \verb|[CLS]| token as the aggregated sequence representation for classification tasks. \citep{devlin-etal-2019-bert} 

\section{Data}
The data set provided consists of 10,000 English Tweets related to COVID-19. Each Tweet is labeled either \verb|INFORMATIVE| or \verb|UNINFORMATIVE|. The Tweets were annotated by three independent annotators with an inter-annotator agreement score of Fleiss’ Kappa at .818. The dataset is partitioned into training, validation, and test sets at a ratio of 7:1:2.  The training set contains 3,303 \verb|INFORMATIVE| and 3,697 \verb|UNINFORMATIVE| Tweets. The validation set contains 472 \verb|INFORMATIVE| and 528 \verb|UNINFORMATIVE| Tweets. The unlabeled Test set contains 944 \verb|INFORMATIVE|, 1,056 \verb|UNINFORMATIVE|, and 10,000 unlabeled Tweets as noise. 

\section{Preprocessing}
To clean the raw Tweets we created a data processing pipeline to: 1) remove non-alphanumeric characters, 2) remove stop-words, 3) convert words to their lemmas, and 4) convert words to lower case. We used the Natural Language Toolkit (NLTK) Python package \citep{bird-loper-2004-nltk} for these methods.

To handle the unique lexicon of Twitter, we implemented additional preprocessing methods to: 1) remove \verb|@USER| tokens, 2) remove \verb|HTTPURL| tokens, 3) remove the \verb|"#"| character, 4) compress repeated characters, and 5) represent emojis as words.

The data set provided replaced URLs and in-Tweet mentions of other users with the \verb|HTTPURL| and \verb|@USER| tokens respectively. Our data-cleaning pipeline removes these tokens when cleaning the data to avoid the models over-fitting to these tokens that do not reflect their original usage within the Tweet. Similarly, we removed the \verb|#| character from Tweets.

We converted emojis into word tokens so that the models would interpret the emojis as words. When using BERT, this allowed the BERT model to generate word embeddings for these emojis

Tweets occasionally contain repeated characters with the purpose of emphasizing a word. For example, \verb|"yesssss"| instead of \verb|"yes"|. In order to consistently capture this kind of emphasis as a separate feature from the original word, we compressed the repetition of a single character into two repetitions of that character.

\section{Methods}
\subsection{Traditional ML models}
Using the \verb|Sklearn| Python package, we generated two separate feature vectors from the preprocessed Tweets. The first method generated feature vectors based on the raw counts of word level unigrams in each Tweet. The second method used TF-IDF to extract word-level unigrams, bigrams, and trigrams features. TF-IDF is a numerical statistic that captures the frequency of a term against the frequency of the documents it appears in. TF-IDF reduces the weight of common words and increases the weight of less frequent words. \citep{Ramos1999} 

We used the \verb|Sklearn| to implement five traditional machine learning models using the features described above: Logistic Regression, Multinomial Naïve-Bayes, Decision Tree, Random Forest, and K-Neighbors \citep{scikit-learn}, using the default hyper-parameters. 

\subsection{Fine-tuned BERT model} \label{BERT}
We first partitioned each Tweet into an array of word tokens. The BERT model requires that each document is the same length, so we padded each array with the embedding \verb|0| so that the length of each entry was 128 tokens. The length of 128 tokens was selected, because the maximum number of words a Tweet could contain within the 255 characters limit is 128 words. We added the \verb|‘[SEP]’| token to the end of each array to denote the end of a sequence. Because we used \verb|BERTForSequenceClassification|, we also added the special \verb|‘[CLS]’| classifier token to the beginning of the array. \citep{devlin-etal-2019-bert}

For our fine-tuning optimizer, we utilized the Adam algorithm with weight decay (AdamW) as introduced in "Decoupled Weight Decay Regularization." \cite{loshchilov2018decoupled} We used the default parameters $\beta_1 = 0.9$, $\beta_2 = 0.999$, and $\verb|epsilon| = .1e-8$. We chose a learning weight of 2e-5 as it offered the lowest training and validation loss when compared to other learning rates between 1e-5 and 1e-4. As fine-tuning BERT required extensive computational resources, we used a Google Colab Research notebook for implementation as it allowed for high-RAM GPU processing. This fine-tuning approach follows the original BERT paper.  \citep{devlin-etal-2019-bert} 

\subsection{Fine-tuned BERT+ model}
In an attempt to capture additional differences in the language of \verb|INFORMATIVE| and \verb|UNINFORMATIVE| Tweets that would not be observed by BERT, we extracted 1024 dimensional embeddings from the the last, non-softmax layer of our fine-tuned BERT and concatenated those with seven Twitter-specific features. We then trained a SVM classifier on these concatenated feature vectors using \verb|Sklearn|'s SVM implementation with default hyper-parameters.

The Twitter-specific features for each Tweet were: 1) Count of the following in the Tweet: \verb|HTTPURL| token, \verb|#|, \verb|@USER| token, and emoji 2) word count, 3) syllable count, and 4) a Boolean specifying whether the Tweet contains profanity. To generate some of these features, we used \verb|PyPI's| \verb|profanity-check|, \verb|syllables|, and \verb|emojis| packages (https://pypi.org/).

\section{Results}
The results reported in subsections \ref{Results-Preprocessing-Features} and \ref{Results-SKLearn} were generated using 8-fold cross validation on the combined train and validation data sets.

\subsection{Preprocessing experiments}\label{Results-Preprocessing-Features}
Our best performing combination of data-cleaning methods, called the Optimal Preprocessor (OP), utilized the Twitter lexicon specific methods we created: 1) \verb|@USER| token removal, 2) \verb|HTTPURL| token removal, 3) \verb|"#"| character removal, 4) repeated character compression, and 5) word representation of emojis. The performance of the OP combined with other preprocessing methods can be seen in Table \ref{preprocess}.

To compare the performances of the \verb|TfidfVectorizer| and \verb|CountVectorizer| for feature extraction, we used the OP for data cleaning and Logistic Regression model for predictions. The \verb|TfidfVectorizer| consistently outperformed the \verb|CountVectorizer| with the average F1-scores of .8422 and .8279 respectively.

\begin{table} [h]
\centering
\begin{tabular}{lrl}
\hline \textbf{Preprocessing Pipeline} & \textbf{F1-score} \\ \hline
Optimal Preprocessor (OP) & \textbf{.8424} \\
OP with Stop-word removal & .8320 \\
OP with Alpha Numeric filter & .8384 \\
OP with forced lowercase & .8351 \\
OP with word lemma conversion & .8337 \\
No preprocessing of data & .8339 \\
\hline
\end{tabular}
\caption{\label{preprocess} Average F1-score from 8-fold cross validation using Logistic Regression model with TF-IDF.}
\end{table} 

\subsection{Traditional ML models}\label{Results-SKLearn}

Using data processed through our OP with features extracted with TF-IDF, we achieved the F1-scores seen in Table \ref{SKLearn-Performance} for our suite of machine learning models. Logistic Regression consistently outperformed the other methods. Table \ref{LogReg-weight} displays selected features with the highest and lowest weights from the Logistic Regression model.

\begin{table}[h]
\centering
\begin{tabular}{lrl}
\hline \textbf{Model/Classifier} & \textbf{F1-score} \\ \hline
Logistic Regression & \textbf{.8422} \\
Multinomial Naïve-Bayes & .8356 \\
Random Forest & .8201 \\
K-Neighbors & .8201\\
Decision Tree & .7313 \\
Baseline Stratified Dummy & .4723 \\
\hline
\end{tabular}
\caption{\label{SKLearn-Performance} Averaged F1-score across K-fold cross validation (k=8) with TF-IDF.}
\end{table} 

\begin{table}[h]
\centering
\begin{tabular}{lrl}
\hline \textbf{Feature (n-gram with n=[1,2,3])} & \textbf{weight} \\ \hline
cases & 5.9113 \\
positive & 5.4668 \\
deaths & 4.9618 \\
died & 4.7764 \\
confirmed & 4.5300 \\
tested & 4.5110 \\
in & 3.9934 \\
positive for & 3.7167 \\
has & 3.6581 \\
tested positive & 3.4429 \\
% ... \\
% china & -1.4569 \\
% can & -1.4591 \\
% trump & -1.5236 \\
% it & -1.5953 \\
% amp & -1.6147 \\
% the & -1.7222 \\
% you & 1.7834 \\
% be & -1.9140 \\
% if & -2.2256 \\
% will & -2.8266 \\
\hline
\end{tabular}
\caption{\label{LogReg-weight} Top 10 Features with weights of the greatest magnitude from the highest performing model (Logistic Regression).}
\end{table} 

\begin{table}
\centering
\begin{tabular}{lrl}
\hline \textbf{BERT Model} & \textbf{F1-Score} \\ \hline
Fine-tuned BERT+ & \textbf{.8713} \\
Fine-tuned BERT & .8701 \\
Pre-trained BERT & .8312 \\
\hline
\end{tabular}
\caption{\label{BERT-Performance} BERT models trained on the train data set and evaluated on the validation data set.}
\end{table} 

\subsection{BERT models}
To generate the results for our BERT models, we fit each BERT model with the train data set and evaluate on the validation data set. The pretrained BERT model yielded a F-1 score of .8312, our fine-tuned BERT model yielded a F-1 score of .8701, and the BERT+ model yielded a further improved F1-score of .8713 (see Table \ref{BERT-Performance}).

\section{Discussion}
\subsection{Preprocessing and feature extraction} 
The method to compress repeated characters in our final data preprocessor might have improved the performance of our model by generating a common word level feature between features that would have been interpreted differently. For example, if one user Tweet, ``good,'' and another user Tweeted, ``gooood,'' these different words would now represent the same feature.

While stop-word removal and word lemma conversion are  commonly in the field of NLP, the presence of the stop words and the complexity of words pre-lemma conversion appeared to help our machine learning models detect stylometric features and improved the performance of our machine learning models. 

For feature extraction, TF-IDF outperformed the word count vectorizer. As TF-IDF decreases the weight of terms that occur frequently across all documents and increases the weight of less common terms, \citep{Ramos1999} it is unsurprising that it outperforms the simple word count feature extraction.

\subsection{Traditional ML models}
With optimized data cleaning and feature extraction, our highest performing traditional models outperformed the base pretrained BERT model. This high performance demonstrates the efficacy of fine tuning the preprocessing steps to achieve competitive performances with these models.

The features in Table \ref{LogReg-weight} depict the features that strongly impact the classification of Tweets in the Logistic Regression model. Based on the description from the WNUT-2020 shared task 2 description of \verb|INFORMATIVE| Tweets as Tweets that, ``provide information about recovered, suspected, confirmed and death cases as well as location or travel history of the cases,'' \citep{covid19tweet} it is unsurprising that word-level features concerning cases, test results, and deaths have large weights.

\subsection{BERT}
Somewhat surprisingly the pre-trained BERT model was outperformed by our logistic regression model for this task. This shows that even for large-scale pre-trained language models such as BERT, task-specific fine-tuning is of utmost importance. 

Though both the fined-tuned BERT and BERT+ models both outperformed the logistic regression model, the difference in performance was not large. (around 3\% boost in performance when using the BERT+ model compared to the logistic regression). We believe this is because BERT is not ideal for classifying noisy Twitter data as it has been trained on well-formed English sentences. This is why several Twitter-specific models have been proposed to deal with noisy Twitter data \cite{tweet2vec, nguyen2020bertweet}.

\section{Conclusion \& Future Work}
In this paper, we have described multiple techniques for automatically identifying and classifying informative COVID-19 Tweets. We have demonstrated the applicability of Logistic Regression with an optimized data cleaning pipeline and TF-IDF for feature extraction for the task of Tweet classification. We have also displayed the higher performance of the BERT+ model. Automated classification of real time data feeds will be important as the COVID-19 pandemic continues to impact the world around us.

For future work, we would to like pre-train a BERT model on a large corpus of Tweets as Twitter’s lexicon and grammatical styling differ from normal usage of the English language. We also want to compare the performance of our fine-tuned BERT model to the performances of other state-of-the-art, pretrained NLP models such as fast.ai’s ULMFIT \citep{howard2018universal} and OpenAI’s GPT2 \citep{noauthororeditor} on this task. We would also like to train a Convolutional Neural Network for Tweet classification. Moreover, as \cite{kim2014convolutional} has demonstrated, Convolutional Neural Network (CNN) trained on top of pre-trained word vectors can achieve state-of-the-art performance for sequence classification. We would like to develop a similar method of utilizing a CNN for the task of Tweet classification.

\bibliographystyle{acl_natbib}
\bibliography{emnlp2020}|

\begin{thebibliography}{13}
\expandafter\ifx\csname natexlab\endcsname\relax\def\natexlab#1{#1}\fi

\bibitem[{Bird and Loper(2004)}]{bird-loper-2004-nltk}
Steven Bird and Edward Loper. 2004.
\newblock {NLTK}: The natural language toolkit.
\newblock In \emph{Proceedings of the {ACL} Interactive Poster and
  Demonstration Sessions}, pages 214--217, Barcelona, Spain. Association for
  Computational Linguistics.

\bibitem[{Devlin et~al.(2019)Devlin, Chang, Lee, and
  Toutanova}]{devlin-etal-2019-bert}
Jacob Devlin, Ming-Wei Chang, Kenton Lee, and Kristina Toutanova. 2019.
\newblock {BERT}: Pre-training of deep bidirectional transformers for language
  understanding.
\newblock In \emph{Proceedings of the 2019 Conference of the North {A}merican
  Chapter of the Association for Computational Linguistics: Human Language
  Technologies, Volume 1 (Long and Short Papers)}, pages 4171--4186,
  Minneapolis, Minnesota. Association for Computational Linguistics.

\bibitem[{Howard and Ruder(2018)}]{howard2018universal}
Jeremy Howard and Sebastian Ruder. 2018.
\newblock Universal language model fine-tuning for text classification.
\newblock In \emph{Proceedings of the 56th Annual Meeting of the Association
  for Computational Linguistics (Volume 1: Long Papers)}, pages 328--339.

\bibitem[{Kim(2014)}]{kim2014convolutional}
Yoon Kim. 2014.
\newblock Convolutional neural networks for sentence classification.
\newblock In \emph{Proceedings of the 2014 Conference on Empirical Methods in
  Natural Language Processing, {EMNLP} 2014, October 25-29, 2014, Doha, Qatar,
  {A} meeting of SIGDAT, a Special Interest Group of the {ACL}}, pages
  1746--1751.

\bibitem[{Liu et~al.(2019)Liu, Ott, Goyal, Du, Joshi, Chen, Levy, Lewis,
  Zettlemoyer, and Stoyanov}]{liu2020roberta}
Yinhan Liu, Myle Ott, Naman Goyal, Jingfei Du, Mandar Joshi, Danqi Chen, Omer
  Levy, Mike Lewis, Luke Zettlemoyer, and Veselin Stoyanov. 2019.
\newblock Roberta: A robustly optimized bert pretraining approach.
\newblock \emph{arXiv preprint arXiv:1907.11692}.

\bibitem[{Loshchilov and Hutter(2018)}]{loshchilov2018decoupled}
Ilya Loshchilov and Frank Hutter. 2018.
\newblock Decoupled weight decay regularization.
\newblock In \emph{International Conference on Learning Representations}.

\bibitem[{Nguyen et~al.(2020{\natexlab{a}})Nguyen, Vu, and
  Nguyen}]{nguyen2020bertweet}
Dat~Quoc Nguyen, Thanh Vu, and Anh~Tuan Nguyen. 2020{\natexlab{a}}.
\newblock Bertweet: A pre-trained language model for english tweets.
\newblock \emph{arXiv preprint arXiv:2005.10200}.

\bibitem[{Nguyen et~al.(2020{\natexlab{b}})Nguyen, Vu, Rahimi, Dao, Nguyen, and
  Doan}]{covid19tweet}
Dat~Quoc Nguyen, Thanh Vu, Afshin Rahimi, Mai~Hoang Dao, Linh~The Nguyen, and
  Long Doan. 2020{\natexlab{b}}.
\newblock {WNUT-2020 Task 2: Identification of Informative COVID-19 English
  Tweets}.
\newblock In \emph{Proceedings of the 6th Workshop on Noisy User-generated
  Text}.

\bibitem[{Pedregosa et~al.(2011)Pedregosa, Varoquaux, Gramfort, Michel,
  Thirion, Grisel, Blondel, Prettenhofer, Weiss, Dubourg et~al.}]{scikit-learn}
Fabian Pedregosa, Ga{\"e}l Varoquaux, Alexandre Gramfort, Vincent Michel,
  Bertrand Thirion, Olivier Grisel, Mathieu Blondel, Peter Prettenhofer, Ron
  Weiss, Vincent Dubourg, et~al. 2011.
\newblock Scikit-learn: Machine learning in python.
\newblock \emph{the Journal of machine Learning research}, 12:2825--2830.

\bibitem[{Radford et~al.()Radford, Wu, Child, Luan, Amodei, and
  Sutskever}]{noauthororeditor}
Alec Radford, Jeffrey Wu, Rewon Child, David Luan, Dario Amodei, and Ilya
  Sutskever.
\newblock Language models are unsupervised multitask learners.

\bibitem[{Ramos et~al.(2003)}]{Ramos1999}
Juan Ramos et~al. 2003.
\newblock Using tf-idf to determine word relevance in document queries.
\newblock In \emph{Proceedings of the first instructional conference on machine
  learning}, volume 242, pages 133--142. New Jersey, USA.

\bibitem[{Vaswani et~al.(2017)Vaswani, Shazeer, Parmar, Uszkoreit, Jones,
  Gomez, Kaiser, and Polosukhin}]{vaswani2017attention}
Ashish Vaswani, Noam Shazeer, Niki Parmar, Jakob Uszkoreit, Llion Jones,
  Aidan~N Gomez, {\L}ukasz Kaiser, and Illia Polosukhin. 2017.
\newblock Attention is all you need.
\newblock \emph{Advances in neural information processing systems},
  30:5998--6008.

\bibitem[{Vosoughi et~al.(2016)Vosoughi, Vijayaraghavan, and Roy}]{tweet2vec}
Soroush Vosoughi, Prashanth Vijayaraghavan, and Deb Roy. 2016.
\newblock Tweet2vec: Learning tweet embeddings using character-level cnn-lstm
  encoder-decoder.
\newblock In \emph{Proceedings of the 39th International ACM SIGIR conference
  on Research and Development in Information Retrieval}, pages 1041--1044.

\end{thebibliography}

\appendix

\end{document}